\documentclass[conference]{IEEEtran}
\usepackage{times}

\usepackage[numbers,sort&compress]{natbib}
\usepackage{multicol}
\usepackage[bookmarks=true]{hyperref}
\usepackage[per-mode=symbol]{siunitx}

\usepackage{booktabs}
\usepackage{lscape}
\usepackage{pifont}
\usepackage{xcolor}
\usepackage{makecell}
\usepackage{soul}
\usepackage{graphicx}
 \usepackage{amsmath}
 \usepackage{amsfonts}
 \usepackage{mathtools}
 \usepackage{bm}

\newcommand{\cmark}{\textcolor{green!80!black}{\ding{51}}}
\newcommand{\xmark}{\textcolor{red}{\ding{55}}}

\hypersetup{
    colorlinks=true,
    linkcolor=blue,
    filecolor=magenta,      
    urlcolor=magenta,
    pdftitle={Overleaf Example},
    pdfpagemode=FullScreen,
    }

\begin{document}

\title{Human-Centered Development of Guide Dog Robots: Quiet and Stable Locomotion Control}

\author{Shangqun Yu$^{*1}$, Hochul Hwang$^{*1}$, \\ Trung M. Dang$^{1}$, Joydeep Biswas$^{2}$, Nicholas A. Giudice$^{3}$, Sunghoon Ivan Lee$^{1}$, Donghyun Kim$^{1}$

}


%

\maketitle
\def\thefootnote{*}\footnotetext{These authors contributed equally to this work}
\def\thefootnote{1}\footnotetext{Manning College of Information and Computer Sciences, University of Massachusetts Amherst. ({\tt\small donghyunkim@cs.umass.edu})}
\def\thefootnote{2}\footnotetext{Department of Computer Science, University of Texas at Austin.
}
\def\thefootnote{3}\footnotetext{School of Computing and Information Science, University of Maine.
}

\begin{abstract}
A quadruped robot is a promising system that can offer assistance comparable to that of dog guides due to its similar form factor. However, various challenges remain in making these robots a reliable option for blind and low-vision (BLV) individuals. Among these challenges, noise and jerky motion during walking are critical drawbacks of existing quadruped robots. While these issues have largely been overlooked in guide dog robot research, our interviews with guide dog handlers and trainers revealed that acoustic and physical disturbances can be particularly disruptive for BLV individuals, who rely heavily on environmental sounds for navigation. To address these issues, we developed a novel walking controller for slow stepping and smooth foot swing/contact while maintaining human walking speed, as well as robust and stable balance control. The controller integrates with a perception system to facilitate locomotion over non-flat terrains, such as stairs. Our controller was extensively tested on the Unitree Go1 robot and, when compared with other control methods, demonstrated significant noise reduction --  half of the default locomotion controller. In this study, we adopt a mixed-methods approach to evaluate its usability with BLV individuals. In our indoor walking experiments, participants compared our controller to the robot's default controller. Results demonstrated superior acceptance of our controller, highlighting its potential to improve the user experience of guide dog robots. Video demonstration (best viewed with audio) available at: \url{https://youtu.be/8-pz_8Hqe6s}.
\end{abstract}


\IEEEpeerreviewmaketitle

\section{Introduction}

\begin{figure} 
    \centering
    \includegraphics[width=\linewidth]{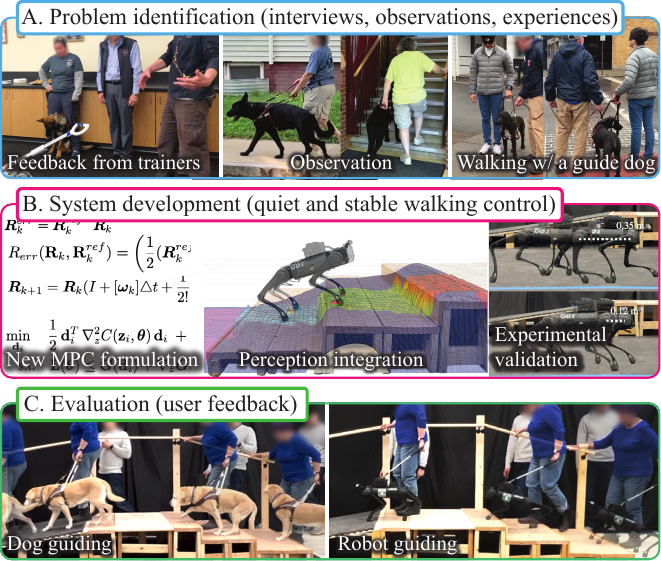}
    \caption{{\bf Human-centered Development of a Guide-dog Robot.} We employed a human-centered approach to develop a guide-dog robot’s locomotion controller, consisting of: 1) an initial exploratory study to identify critical unmet needs, 2) system (technology) development, and 3) user evaluation. One of our major contributions is the adoption of this complete, user-centric development process to advance the robot controller for blind and low-vision individuals.}  
    \label{fig:hcd}
\end{figure}

Guide dogs have long been recognized as an effective mobility aid, enhancing blind and low-vision (BLV) individuals’ navigation capabilities, confidence, and independence~\cite{miner2001experience, whitmarsh2005benefits}. Despite these benefits, only a small fraction of the 250 million BLV individuals worldwide~\cite{ackland2017world} work with guide dogs (unofficial estimates suggest around 2\%\cite{howmany-guidedog,CANADA}), primarily because of their limited availability. Each guide dog can cost as much as \$40,000 to train for up to two years at specialized guide dog schools\cite{pit23,geb23,cnc23}. Moreover, deciding to work with a guide dog is a multifaceted process for BLV individuals, involving health~\cite{GDoA, GDF} and financial eligibility~\cite{pit23}, ongoing ownership responsibilities, and the emotional challenges of managing a guide dog’s retirement~\cite{schneider2005practice}.

In response, researchers have been exploring robotic solutions since 1976~\cite{tachi1978study}, investigating prototypes ranging from smart canes~\cite{slade2021multimodal} and suitcase-shaped devices~\cite{guerreiro2019cabot, kuribayashi2023pathfinder} to flying drones~\cite{avila2017dronenavigator, al2016exploring, tan2021flying}. These robotic aids offer greater scalability, cost-effectiveness, and consistent reliability compared to animal guides. Recent advances in quadruped locomotion control and the availability of increasingly affordable quadruped robots have prompted investigations into robots with a dog–like form factor. However, beyond technical hurdles, numerous challenges persist in developing assistive systems for BLV individuals, including the complicated tasks of understanding users’ needs and identifying critical gaps~\cite{hwang2024towards}.

While prior human-centered studies have uncovered various important aspects, most have focused on wheeled robots and navigation perspectives~\cite{guerreiro2019cabot, zhang2023follower, kuribayashi2023pathfinder, kayukawa2019bbeep,liu2024dragon,nanavati2018coupled}. Although several studies have suggested how to develop quadruped robots for BLV users \cite{cai2024navigating, kim2025understanding, wang2023can, hwang2023system, hwang2024towards}, the question of `how the robot should walk' remains largely underexamined. In particular, Wang et al. emphasized the importance of noise suppression based on BLV individuals’ responses to quadruped robots in comparison to wheeled systems, yet neither the criticality of low noise levels nor clear thresholds for acceptability were presented~\cite{wang2023can}. Kim et al. compared user preferences between a default controller and an RL-based controller, it did not present any measured noise levels or quantitative control performance metrics, limiting its utility as a guideline for controller design~\cite{kim2025understanding}. Moreover, existing studies rarely offer a holistic perspective on related factors such as walking speed and robustness against pulling forces -- issues that must be investigated in tandem, given that lowering noise often involves slowing the stepping frequency and potentially compromising both pace and stability. Equally important, detailed accounts of how guide dogs and handlers navigate uneven terrains (e.g., bumps, curbs, or stairs) remain scarce, further underscoring the need for research on how a quadruped robot should walk in real-world assistive contexts. 

To identify technical gaps, set appropriate goals, and make informed development decisions, we conducted comprehensive exploratory studies. These studies included interviews with seven guide dog handlers and one trainer, demonstrations of a quadruped robot to gather feedback from stakeholders, and a two-and-a-half-hour blindfolded walking experience under expert supervision (Fig.~\ref{fig:hcd}). We found that BLV individuals often rely on subtle auditory cues for navigation, making the default robot controller’s noise levels difficult to accept. We also discovered that their walking speed can be high -- sometimes exceeding the typical pace of a sighted person -- when they walk with a guide dog. Additionally, we learned how guide dogs behave when upcoming stairs or bumps are detected. These findings provide invaluable insights for designing a suitable locomotion controller for our guide dog robot.

Based on our exploratory studies, we prioritized the reduction of robot noise. When a quadruped moves, two primary sources of noise arise: (1) impact noise from foot-ground contact and (2) collision noise between linkages and gears due to mechanical backlash. While software solutions cannot entirely eliminate backlash noise, lowering the gait frequency and ensuring gentle touchdowns can effectively mitigate both backlash-induced and impact noise. One challenge is maintaining stability at slower gait frequencies, which necessitates robust balance control, as slower stepping generally reduces stability compared to more frequent stepping.

To reduce noise, Kim et al. developed an RL-based locomotion controller~\cite{kim2025understanding}. In our case, we chose to improve the model predictive control (MPC)-based locomotion controller rather than adopt RL-based methods due to three reasons: (1) although RL-based locomotion control has made significant progress, most studies prioritize agility~\cite{10610200, zhuang2023robot, 10678805} rather than the gentle walking we require. Conversely, MPC-based controllers~\cite{bledt2019implementing} have demonstrated promising balance capability at slower stepping rates; (2) several existing RL-based controllers that achieve quieter walking~\cite{kim2025understanding, valsecchi2024accurate} do not integrate terrain perception. Incorporating vision while maintaining quiet steps can pose substantial challenges: domain randomization, a standard technique to handle sim-to-real gaps, typically randomizes camera placement and thus introduces uncertainty in terrain height. Our experiments show that even a few centimeters of error in height measurements can result in leg stumping at touchdown, implying substantial difficulty in obtaining gentle touch-down using RL-based methods; and (3) demonstrating specific robot behaviors -- such as placing the front feet on the first staircase step to inform the handler of an upcoming incline -- is easier with an optimization-based controller. This is not to imply the limitations of either learning-based or optimization-based approaches; with further research, both methods may achieve comparable outcomes. In the current state, optimization-based approaches demonstrate promising capabilities and offer convenient features, such as direct management of swing leg motion.

Despite the promise of optimization-based control for the walking behavior we envision, no existing MPC formulation~\cite{kim2019highly, bledt2019implementing, grandia2023perceptive} explicitly addresses a noise suppression issue alongside vision-based walking. Through extensive comparison of various formulations, we concluded: (1) accurately modeling angular motion without simplification is beneficial, (2) a high update frequency is essential, and (3) separately managing swing foot control is especially advantageous for non-flat terrain. Considering the limited computational resources and the lightweight nature of our quadruped’s legs, we developed a walking controller comprising a real-time sequential quadratic programming (SQP)-based Nonlinear Model Predictive Control (NMPC) and a whole-body impulse controller (WBIC)~\cite{kim2019highly}. In our formulation, we preserve the nonlinear angular dynamics via an $\mathcal{SO}(3)$ representation similar to \cite{ding2021representation} and achieve a high update frequency by using real-time SQP~\cite{diehl2005real}. 

Our new controller accomplishes noise reduction of up to 10~dB compared to the robot’s default controllers. Compared with the quadruped robot walking presented in \cite{valsecchi2024accurate} and wheeled systems tested in \cite{wang2023can}, our controller exhibits lower noise levels. To evaluate these improvements from the user’s perspective, we recruited four guide dog handlers to compare two different controllers in flat, stair, and ramp conditions, assessing noise, usability, and workload. Their feedback emphasized a noticeably lower noise level, which they highly valued. All four BLV participants reported that our locomotion controller significantly reduced noise, offered a perceived workload comparable to a guide dog, increased satisfaction, and facilitated comfortable stair climbing. Also, to our knowledge, this constitutes the first systematic investigation of BLV individuals’ interactions with quadruped robots in challenging contexts such as stair climbing, providing critical insights for harnessing legged platforms as mobility aids. Consistent with common usage, we will use the term \emph{guide dog} instead of the formal \emph{dog guide} throughout this paper.

The primary contributions of this paper are threefold:  

\begin{enumerate}
    \item We identify robot control elements required to enhance user experience during walking and stair climbing with a guide dog robot, grounded in semi-structured interviews, observational, and participatory sessions with stakeholders.
    \item We have developed a locomotion control framework that can substantially reduce noise, achieving up to a 50\% reduction in noise level compared to existing controllers. Our framework also incorporates perception-based stair climbing to enhance overall usability across varied terrains.
    \item We conduct a comprehensive, user-centric evaluation -- incorporating both qualitative and quantitative metrics -- that assesses noise, usability, and workload with BLV people. 
\end{enumerate}




\section{Related works}

Mobile robots for BLV individuals have been studied over 40 years, beginning with wheeled systems~\cite{Tachi, meldog, tobita2018structure, guerreiro2019cabot, tobita2017examination, kulyukin2004robotic, megalingam2019autonomous, nanavati2018coupled, galatas2011eyedog, kayukawa2019bbeep} and more recently moving to quadruped robots~\cite{xiao2021robotic, chen2022quadruped, hamed2019hierarchical, NSK, due2023walk, cai2024navigating}. However, despite the long history of assistive devices for individuals with blindness or low vision (BLV), critical gaps persist between the functionalities of existing systems and the needs of users~\cite{hwang2024towards}. Here, we provide a brief summary of existing systems, categorizing them into several broad groups:

\emph{Handheld systems} like smart canes~\cite{smartcane, slade2021multimodal, ye2016co, ulrich2001guidecane, takizawa2015kinect, takizawa2012kinect, faria2010electronic, saaid2016smart, wewalk}, or wearable devices~\cite{belt, headmount, zeng2017camera, katzschmann2018safe, li2016isana, rodriguez2012assisting, strumillo2018different, dakopoulos2009wearable}, offer convenient access to navigation assistance at reasonable price points. However, these devices face challenges as they provide navigation cues through acoustic, vibrotactile, or gentle force feedback, which often leads to reduced feedback effectiveness. This issue arises because users may become numb due to repeated tactile signals or experience distraction and mental fatigue, particularly for BLV users who rely heavily on environmental sounds~\cite{strothotte1996development}. Although some recently introduced systems, such as Glide~\cite{glidance}, attempt to address these limitations by better balancing these input modalities, the fundamental constraints of pulling or pushing forces limit their ability to respond to emerging risks. For example, unlike guide dogs trained to pull or push their handlers away from danger~\cite{guide_dog_documentary}, these devices can only provide alarms (e.g., a beep or vibration), leaving users to decide their next move. Moreover, limitations in sensing and computation -- imposed by the small size and lightweight design of these devices -- further restrict their ability to deliver comprehensive navigation assistance features. 

On the other hand, \emph{wheeled mobile robots}, which provide direct physical guidance (e.g., pulling or pushing), overcome some limitations of handheld systems~\cite{Tachi, chuang2018deep, meldog, tobita2018structure, guerreiro2019cabot, tobita2017examination, kulyukin2004robotic, megalingam2019autonomous, nanavati2018coupled, galatas2011eyedog, kayukawa2019bbeep}. While some wheeled systems, such as CaBot~\cite{guerreiro2019cabot} and Lysa~\cite{lysa}, have demonstrated strong indoor navigation capabilities, their designs inherently limit them to smooth surfaces and controlled environments. Their inability to navigate uneven terrains, such as stairs or curbs commonly found in real-world outdoor settings, restricts their usefulness in the environments where BLV individuals need the most support.

{\emph{Legged robots}}, with their terrain adaptability and ability to provide physical guidance, can address these limitations by mimicking how guide dogs assist their handlers. While promising, many quadruped robots remain inadequately designed, mainly due to the lack of understanding of how guide dogs and their handlers work in real-life situations. For instance, many existing systems employed a soft leash~\cite{xiao2021robotic, chen2022quadruped, kim2023train, glasgow_dog}, which is problematic because handlers should use a rigid harness handle to perceive immediate feedback from the guide dog’s motion~\cite{guerreiro2019cabot}. In other studies~\cite{hamed2019hierarchical, cai2024navigating, jiaotong6}, the handler was positioned behind the robot during navigation, whereas guide-dog handlers should walk beside the dog for safety. Some studies address technical problems that are less critical or unimportant for assisting BLV navigation. For example, \cite{hamed2019hierarchical} focused on improving the locomotion stability of quadruped robots to withstand the pulling forces from a handler. However, the study measuring the pulling forces of guide dogs showed that the robot’s default controller is already sufficient to sustain such forces~\cite{hwang2023system}. 

To minimize such errors, we adopted a human-centered approach to guide dog robot development. This approach emphasizes understanding the real-world needs of BLV individuals and the dynamics of interactions between guide dogs and their handlers. In this paper, we focus on the locomotion of quadruped robots, which is a fundamental feature that shapes the walking experience of BLV individuals when navigating alongside guide-dog robots.

\section{Lessons learned from Stakeholders} 
Our exploratory study aimed to identify technical gaps in improving the user experience for BLV individuals when walking with a guide dog robot. To achieve this, we conducted semi-structured interviews and observation sessions with guide dog trainers and handlers. Furthermore, we also participated in blindfolded walking with guide dogs under the trainer's supervision to obtain a lively experience of how animal guide dogs assist their handlers. 



\subsection{Exploratory User Study Design}
\subsubsection{Participants}
 We recruited four guide dog handlers and two professional guide dog trainers. Inclusion criteria for handlers included (1) a visual acuity of 20/200 or worse, as defined by the Social Security Act \S 1614~\cite{ssa20}, and (2) at least six months of guide dog experience. Trainers were required to have a minimum of five years of experience in guide dog training. Demographic details for both groups are presented in Table~\ref{tab:exploratory}.

\subsubsection{Procedure}
The interview questions were co-designed with a guide dog handler and trainer which focused on the interactions within the handler-guide dog team and the functionalities of animal guide dogs. During observation sessions, participants navigated familiar routes with their guide dogs covering both indoor and outdoor environments, providing live demonstrations of real-world challenges and interactions. This included climbing up stairs of a building and ascending staircases to ride a bus. Trainers also demonstrated specialized training techniques, emphasizing how handlers work with their dogs in daily life.



\begin{table}
  \caption{Interview and Observation Session Participant Demographics}
  \label{tab:exploratory}
  \begin{tabular}{cccccc}
    \toprule
    \textbf{ID} & \textbf{Age} & \textbf{Gender} & \textbf{Vision Level} & \textbf{Mobility Aid} & \textbf{\makecell[c]{Experience\\(GD yrs$^{*}$)}}\\
    \midrule
    H01 & 61 & F & Totally blind & Guide dog & 36 \\
    H02 & 69 & F & Legally blind & Guide dog & 30 \\
    H03 & 63 & M & Totally blind & Guide dog & 11 \\
    H04 & 69 & M & Totally blind & Guide dog & 30 \\
    H05 & 69 & M & Totally blind & Guide dog & 53 \\
    H06 & 59 & M & Totally blind & Guide dog & 21 \\
    H07 & 64 & F & Totally blind & Guide dog & 6 \\
    T01 & 58 & M &- &- & 35\\

    
    \bottomrule
\end{tabular}
\footnotesize{\\$^{*}$ Indicates the total years participants spent working with guide dogs as handlers (H) and training guide dogs as trainers (T).} 
\footnotesize{\\$^{\dag}$ Denotes the experience of walking with a guide dog robot prior to this study.}\\
\end{table}

\begin{figure*}
    \centering
    \includegraphics[width=\textwidth]{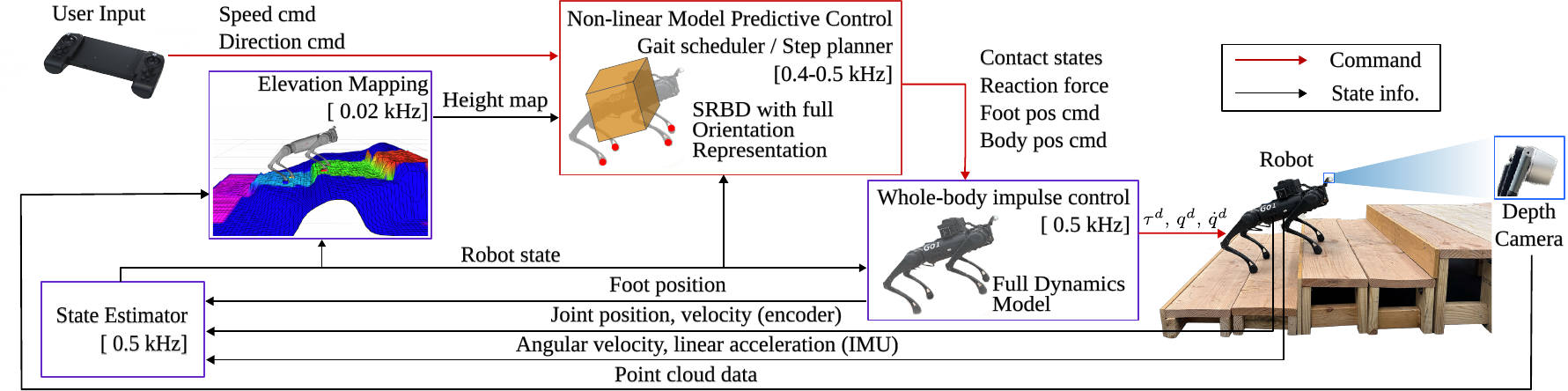}
    \caption{{\bf Overall Control Framework.} Our guide-dog controller uses an NMPC+WBIC architecture. The NMPC incorporates a single-rigid-body model with full orientation dynamics to enable better orientation control and computes the required reaction forces, which are then passed to the WBIC. The WBIC generates feedforward torques, as well as desired joint positions and velocities, which are subsequently fed into the robot’s onboard motor controller. For stair climbing, the robot is equipped with a RealSense D435 camera, and the captured point clouds are processed into a height map using Elevation Mapping. }  
    \label{fig:GuideDogSystem}
\end{figure*}


\subsection{Findings}
\subsubsection{Noise: Can we use the robot's off-the-shelf locomotion controller?} 
Multiple handlers emphasized the importance of auditory cues in their everyday mobility. They rely on subtle environmental sounds –- like approaching vehicles or echoic feedback from walls –- to interpret their surroundings.
\begin{quote}
{“Some people don't believe this, but I do believe that we have this ability to hear the echos, or a different, a deadening sound of the heels on your shoes.”} - \textbf{H03}
\end{quote}

Off-the-shelf quadruped robot controllers, however, can generate substantial noise, which risks masking essential audio information. Based on our measurements, the noise of the Unitree Go1’s trot gait is similar to a vacuum cleaner, and this can be worsened in indoor environments, as H03 mentioned, “The noise is a little distracting. And the banging in the house. [...] You’re not gonna sneak up on somebody that’s for sure.”

Even low-level buzzing, repetitive clicks, or whirring motors can disrupt the user’s natural orientation process or create stress during navigation, which also aligns with prior studies that had blind people walk with quadruped robots~\cite{kim2025understanding,wang2023can}. As a result, participants expressed concern that high noise levels might compromise situational awareness and confidence. 

In conclusion, we found that noise suppression is more than a mere preference; in fact, the required reduction in noise levels is quite substantial, and achieving near-complete silence, if possible, would be advantageous. Consequently, we set our subsequent research goal to minimize the robot’s noise as much as feasible.



\subsubsection{Walking speed: Do blind people walk slower to be careful?}
Contrary to common assumptions, the handlers we observed did not slow their pace for safety. Rather, they aimed to maintain a preferable walking speed. Moreover, some handlers walked with their guide dogs at a walking speed comparable to that of sighted walkers, as H01 mentioned after the observation session when the researcher asked if the handler preferred such high speed walking: “I walk fast with him. [...] Well, I like moving that fast.” Consequently, the guide dog robot must accommodate a range of walking speeds, including those that may exceed average walking speed, while ensuring both safety and user comfort. 


\subsubsection{Stair Climbing: How do guide dogs assist stair climbing?}

People perceive when the staircase starts before climbing up stairs. How do BLV individuals obtain such cues? Based on the observation of H01 climbing up stairs and H02 climbing up a bus with their guide dogs, we noticed certain promised sequences for safe stair climbing. Handlers perceived staircases through their harness, delivering the subtle movement of their dogs. As H01 explained, her guide dog stops after stepping on the first stair, allowing her to notice the presence of stairs ahead by detecting the tilt of the handle.

\begin{quote}
{“He stopped on the first step when he was on it. But I told him to go. But I need that pause just to hit it. What I did was I kicked it with my toe and I knew it was the stairs. I knew they were upstairs because he put his paws up. Versus down, I could tell if it was up or down on that first step.”} - \textbf{H01}
\end{quote}


In summary, our findings highlight the critical need for a locomotion controller that can achieve high walking speeds (\SIrange{0.6}{1.2}{\meter\per\second}) and stair-climbing capabilities without producing disturbing noise or jerky movements. These insights guided the development of a user-centric locomotion controller tailored to BLV individuals.

\section{Control framework} 
\label{sec:method}

\subsection{Real-time Nonlinear MPC}

To develop a noise-suppressed guide-dog locomotion controller, our primary goal was to reduce the stepping frequency and ensure gentle foot contacts by employing a combined NMPC+WBIC architecture (fig. \ref{fig:GuideDogSystem}). The NMPC model includes a full orientation representation and is solved via a single-iteration Sequential Quadratic Programming (SQP) framework, where the inner quadratic program is efficiently handled using the Alternating Direction Method of Multipliers (ADMM). This approach allows for an updated frequency of up to 500~Hz, thereby minimizing latency and enhancing overall robot stability. Meanwhile, Whole Body Impulse Control (WBIC) tracks the reaction forces proposed by NMPC and simultaneously manages foot-swing trajectories, contributing to smoother, quieter locomotion.

\subsubsection{Nonlinear Model Predictive Control (NMPC)}

To accurately represent angular motion, we used $\mathcal{SO}(3)$ representation instead of Euler angle similar to the state formulation presented in \cite{10769916}. The state vector for the MPC is defined as:
  \begin{equation}
     \mathbf{x} = [\mathbf{p^\top} \quad \mathbf{R_{vec}} \quad \mathbf{v^\top} \quad \bm{\omega^\top}]^{\top},
 \end{equation}
where $\mathbf{p} \in \mathbb{R}^3$, $\mathbf{v} \in \mathbb{R}^3$, $\mathbf{R_{vec}} \in \mathbb{R}^9$, and $\bm{\omega} \in \mathbb{R}^3$ are position, velocity, vectorized orientation matrix and angular velocity of the single rigid body. 
The cost function to be minimized is:

\begin{equation*}
\min_{\mathbf{p}_k,\mathbf{f}_k} \quad  \sum_{k=1}^{N} \ ref^{err} (\mathbf{x}_k, \mathbf{x}^{ref}_k) +\Vert \mathbf{f}_k - \mathbf{f}^{ref}_k \Vert^2_{\bm{Q_f}} 
\end{equation*}
where
\begin{align*}
ref^{err} = \Vert \mathbf{p}_k - \mathbf{p}^{ref}_k \Vert^2_{\bm{Q_p}} + \Vert \mathbf{v}_k - \mathbf{v}^{ref}_k \Vert^2_{\bm{Q_v}} \\ + \ \Vert \bm{\omega}_k - \bm{\omega}^{ref}_k \Vert^2_{\bm{Q_\omega}} + \Vert R_{err} (\mathbf{R}_{k}, \mathbf{R}_{k}^{ref}  ) \Vert^2_{\bm{Q_R}}
\end{align*}

\begin{align*}
\textrm{s.t.} \quad & \mathbf{x}_{k+1} = f_k^d( \mathbf{x}_k, \mathbf{f}_k),  &\textrm{(dynamics)} \\[1mm]
  & -\mu f_{k,i,z} \leq f_{k,i,x} \leq \mu f_{k,i,z},   \\[1mm]
  & -\mu f_{k,i,z} \leq f_{k,i,y} \leq \mu f_{k,i,z},  &\textrm{(friction cone)}  \\[1mm]
  & 0 \leq f_{k,i,z} \leq f_{max}, & \textrm{(reaction force)}\\[1mm]
  & f_{k,i,z} (1 - c_{k,i}) = 0, & \textrm{(contact scheduling)}
\end{align*}
where $\mathbf{x}^{ref}_k \in \mathbb{R}^{18}$ and $\mathbf{f}^{ref}_k \in \mathbb{R}^{12}$ are the reference for the state and reaction force, respectively. $\bm{Q_p}$,  $\bm{Q_v}$,  $\bm{Q_\omega}$,  $\bm{Q_R}$ and $\bm{Q_f}$ are the corresponding diagonal weight matrices. $f_k^d$ is the forward dynamics for a single rigid body, and $c_{k}$ presents a contact schedule specified by a user. The reference reaction forces $\mathbf{f}^{ref}_k$ are simply set as vertical reaction forces that support the robot’s weight according to the contact schedule. $\mathbf{x}^{ref}_k$ are computed by integrating the current state forward given a commanded base velocity. 

Except for the orientation, reference errors are L2 norm of the difference between the state and reference. The orientation error $\bm{R}^{err}_k$ is given as 
\begin{equation}
\bm{R}^{err}_k = {\bm{R}_k^{ref}}^\top \bm{R}_k 
 \end{equation}
where $\bm{R}^{ref}_k$ and $\bm{R}_k$ is the desired and actual orientation matrix at $k$ step. To get vectorized orientation error, we transform $\bm{R}^{err}_k$ to $\mathfrak{so}(3)$ using 
  \begin{equation}
    R_{err} (\mathbf{R}_{k}, \mathbf{R}_{k}^{ref}  ) =  \left(\frac{1}{2} ({\bm{R}_k^{ref}}^\top \bm{R}_k  - {\bm{R}_k^\top \bm{R}_k^{ref}})\right)^{\vee},
 \end{equation}
where $( \cdot)^\vee : \mathfrak{so}(3) \rightarrow \mathbb{R}^3$ is the inverse of the skew function. The only simplification here is $\theta^{err}_k \approx \sin{\theta^{err}_k}$ based on the small orientation error assumption.


\begin{figure*}
    \includegraphics[width=\linewidth]{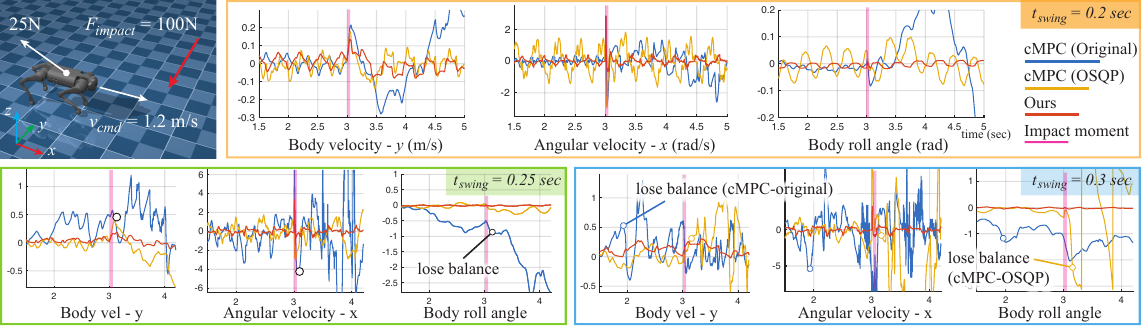}
    \caption{{\bf Comparison of Our MPC and Convex MPC.} We compare our controller with the convex MPC + WBIC framework presented in \cite{kim2019highly} to evaluate the improved balance stability resulting from our new MPC formulation. In the Mujoco simulation, we command the robot to move forward at $1.2~\si{\meter\per\second}$ while applying a $25~\si{\newton}$ pulling force to simulate the force exerted by a handler. During walking, we apply an impulse force of $100~\si{\newton}$ for 0.01 seconds to assess the controllers' robustness. The tests were performed at three different gait speeds (i.e., swing times of 0.2, 0.25, and 0.3 seconds). The results show that, in both the original implementation and the new implementation using OSQP, convex MPC exhibits limited capability in sustaining balanced walking compared to ours.}
    \label{fig:sim_results}
\end{figure*}

 The base orientation update is formulated by the 3rd order of the Taylor expansion to minimize the approximation error in matrix exponential: 
  \begin{equation}
     \bm{R}_{k+1} = \bm{R}_k ( I + [\bm{\omega}_k] \triangle t + \frac{1}{2!} [\bm{\omega}_k]^2 \triangle t^2 + \frac{1}{3!} [\bm{\omega}_k]^3 \triangle t^3 ).
 \end{equation}

While our formula accurately represents angular motion dynamics, it introduces nonlinearity into the MPC formulation, resulting in longer computation times. However, to enhance the robot's stability at low gait frequencies, a higher update rate for the MPC is essential. To achieve frequent updates without simplifying the formulations, we adopted the real-time Sequential Quadratic Programming (SQP) approach, initially proposed in \cite{diehl2005real} and recently applied in robot control \cite{grandia2022perceptive, khazoom2024tailoring}. The primary idea involves using intermittent solutions of SQP for control without waiting for full convergence. Specifically, the NMPC problem is approximated by the following quadratic programming (QP): 
\begin{equation} \label{eq:qp_appr}
\begin{split}
&\min_{\mathbf{d}_i} \quad \frac{1}{2}\, \mathbf{d}_i^T \,\nabla_z^2 C(\mathbf{z}_i,\bm{\theta})\,\mathbf{d}_i
   \;+\;\nabla_z C(\mathbf{z}_i,\bm{\theta})^T \mathbf{d}_i,
\\
&\quad \textrm{s.t.} \quad  \mathbf{L}(\bm{\theta}) \leq \bm{G}(\mathbf{d}_i) + \nabla_z \bm{G} (\mathbf{z}_i,\bm{\theta})\,d_i    \leq \mathbf{U}(\bm{\theta}),
\end{split}
\end{equation}
which are derived from the original equations:
\begin{equation}
\begin{split}
\min_{\mathbf{z}} \quad  &C(\mathbf{z}, \bm{\theta}) \\
\textrm{s.t.} \quad  \mathbf{L}(\bm{\theta}) \leq &\bm{G}(\mathbf{z}, \bm{\theta}) \leq \mathbf{U}(\bm{\theta}),
\end{split}    
\end{equation}
where $\mathbf{z}$ and $\bm{\theta}$ is the decision variable and the parameter vector. $\bm{G}(\mathbf{z})$ is the constraints with the lower bound $\mathbf{L}(\bm{\theta})$ and upper bound $\mathbf{U}(\bm{\theta})$.

In approximated QP~\eqref{eq:qp_appr}, $\mathbf{d}_i$ is the step direction, $\nabla_z^2 C(\cdot)$ is the Gauss-Newton Hessian of the cost function $C(\cdot)$. $\nabla_z C(\cdot)$ is the gradient of $C(\cdot)$ and $\nabla_z \bm{G}(\cdot) $ is the Jacobian of $\bm{G}(\cdot)$. The cost function is essentially a second-order Taylor approximation of the original cost while the constraint is a linear approximation of the original constraint. Once QP finds $\mathbf{d}_i$, the decision variable is updated as:
\begin{gather*}
    \bm{z}_{i+1} = \bm{z}_i + \alpha_i \bm{d}_i,
\end{gather*}
where $\alpha$ is a step size found based on a line search algorithm from \cite{10138309}. In a standard SQP setup, this process should be repeated multiple times until $\bm{z}$ converges to a local minimum. However, in a real-time SQP setup, the first timestamp of $\bm{z}$ (noting that $\bm{z}$ includes a long-horizon solution) is used for control without waiting for the convergence. As the robot moves forward, SQP solves a new problem similar to the previous one; thus, we can use the previous premature solution as a warm-starting point for the new problem. Surprisingly, this strategy works well in real-time control with proper configuration, as extensively analyzed in \cite{khazoom2024tailoring} and also empirically observed in this study. For the final stage, Whole Body Impedance Control (WBIC)~\cite{kim2019highly} is employed to compute the feedforward torque $\bm{\tau}_{ff}$, desired joint positions $\bm{q}_{des}$, and desired joint velocities $\bm{\dot{q}}_{des}$.

\subsubsection{Other details of motion control} 
Except for switching convex MPC with our new real-time SQP formulation, other parts are similar to the implementation in \cite{kim2019highly}, such as using the Raibert heuristic and capture point-based landing location selection, WBIC task and constraints setup, and weight parameters. Several adjustments were made to enable smoother touchdown and noise suppression. While tuning the low-level joint position and derivative feedback controller, we found that a higher value of the derivative gain, $\mathbf{K}_d$, tends to induce more motor winding noise. Consequently, for flat terrain walking, $\mathbf{K}_d$ is set to 0.5, which is much smaller than the nominal value like 1 or 2. for all joints to minimize noise and improve compliance with touchdown impact. In the case of stair climbing, $\mathbf{K}_d$ is adjusted to 2, 1, and 1 for the hip abduction, hip pitch, and knee joints, respectively, to enhance foot tracking performance, as landing location accuracy is more critical than in flat terrain walking. Swing trajectory timing is also slightly adjusted to slow down touchdown velocity. In the original implementation, lift-off and landing motions are evenly distributed over the swing time. Instead, we allocate a longer duration to the downward swing (65\% of the swing time), thereby reducing impact noise upon ground contact.


\subsection{Stair climbing}
\label{subsec:stair}
To traverse common non-flat terrains, including stairs, we integrated stair-climbing capabilities into the controller. Similar to \cite{kim2020vision}, we use a 2.5D height map for terrain representation and adjust the landing location and height based on the map information. Specifically, to prevent the robot from stepping on stair edges, we modify the landing location within a local region where the normal vectors are nearly vertical.

For guiding stair climbing for BLV individuals, guide dogs are trained to pause at the beginning of a staircase, allowing blind individuals to prepare for climbing. This pre-stopping functionality is integrated into our guide dog controller by implementing a feature that halts the robot when its front two feet make contact with the first staircase step. Once the user is ready, the robot proceeds to climb the stairs.






\section{Controller Evaluation} 
\label{sec:experiment}
To analyze the performance of the proposed controller compared to other methods, we perform extensive tests both in simulation and on robot hardware. To implement our MPC, we first compute the gradients and Hessians of the cost function and constraints (presented in Eq.~\eqref{eq:qp_appr}) via automatic differentiation using CasADi \cite{andersson2019casadi}. In every MPC iteration, OSQP \cite{osqp} is used to solve the QP problem, which is equivalent to a single SQP iteration. Unlike common implementations that maintain a fixed update frequency, our MPC, running on an independent thread, starts a new iteration as soon as the previous computation is completed, thereby minimizing delays and enhancing balance control. Additionally, we warm-start each iteration with the solution from the previous cycle, leveraging the fact that the problem formulation changes only marginally over time. A single QP solve usually takes around 2 ms, leading to an MPC running frequency of 400 Hz – 500 Hz. In the case of WBIC, it runs within the main control loop at a fixed frequency of 500 Hz. The final command computed by WBIC, joint torque, position, and velocity commands, are sent to the robot's joint-level controller using the Unitree SDK.

\begin{figure*} 
    \centering
    \includegraphics[width=\linewidth]{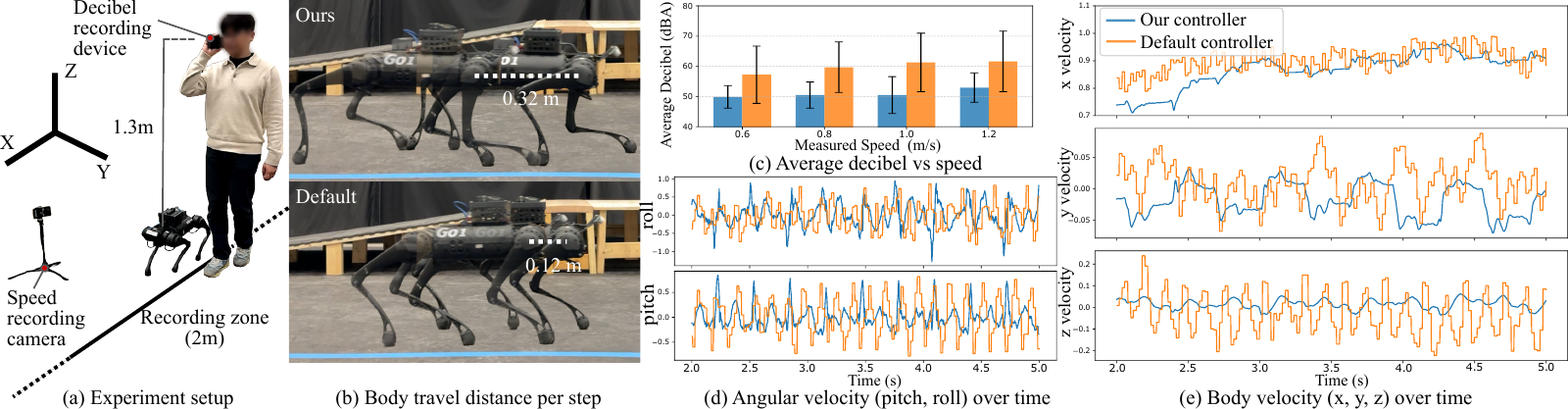}
    \caption{{\bf Hardware Experiment Results.} (a) To accurately measure the noise BLV individuals hear during walking, a person records the decibel value at his ear level while following the robot. (b) Both controllers (Our vs Default) are traveling at 1 m/s. The slower gait requires our controller to travel more distance within a single step. (c) On average, our controller has achieve a 10 decibel reduction. (d) We observed that angular velocity amplitudes of each controller are similar or even less sometimes. This shows our approach maintains comparable or better orientation control while having slower gait frequency. (e) Our controller shows better performance in maintaining the body velocity.}  
    \label{fig:hardware_exp}
\end{figure*}

\subsection{Comparison of Model Predictive Controllers}
To validate the effectiveness of incorporating accurate angular motion dynamics and to demonstrate the improved stability and robustness in walking due to our MPC formulation, we compare our approach with the convex MPC + WBIC controller (referred to as cMPC from now on), introduced in \cite{kim2019highly}. Here, Mujoco simulation~\cite{todorov2012mujoco} is utilized to create an environment that enables quantitative analysis by eliminating uncertainties in modeling and sensing.

In the tests, we set the desired forward speed to $1.2\si{\meter\per\second}$ while applying a continuous dragging force ($25\si{\newton}$ directed backward and vertically at $45\si{\degree}$) to mimic the pulling force from a handler. While \cite{hwang2023system} reported that $34.32\si{\newton}$ was observed during blindfolded walking with a guide dog, our experience and feedback from a guide dog trainer indicated that this force is excessive -- the recommended handling method involves gently holding the harness to detect subtle motions of the dog. Thus, we set the dragging force to $25\si{\newton}$, which remains high but is not excessively strong. To test robustness, we apply an additional external push force of $100\si{\newton}$ to the robot mid-walk for a short duration (0.01\si{\second}).

While cMPC accomplished high-speed running of a quadruped robot~\cite{kim2019highly}, it exhibits immediate limitations in stabilizing a trot gait with slow stepping, as evidenced by the loss of balance under disturbances when the swing time is set to 0.25~\si{\second}. When the swing time is increased to 0.3~\si{\second}, cMPC fails to achieve stable walking, as shown in Fig.~\ref{fig:sim_results}. This instability is partially due to the original implementation of cMPC~\cite{BiomimeticsSoftware}. The original implementation uses 1) a fixed update frequency that slows down as the MPC time step increases, and 2) qpOASES~\cite{Ferreau2014} with dense matrix calculations instead of leveraging the problem’s sparsity. These factors negatively impact stability when the swing time increases, as they require either increasing the time step or extending the horizon, which reduces the update frequency or significantly complicates the problem's complexity, respectively.

For a fair comparison, we reimplemented cMPC in our MPC setup using a flexible update frequency, OSQP, and CasADi (referred to as cMPC (OSQP)). Thanks to the improved computational setup, cMPC (OSQP) exhibits better stability at a slow-paced gait and greater robustness against external impacts. However, its balance control remains noticeably less stable than ours, as observed in the angular velocity and orientation plots in Fig.~\ref{fig:sim_results}. In contrast, the proposed controller (ours) successfully manages balance control under continuous dragging forces and sudden impact disturbances, demonstrating superior locomotion performance.

For implementation details, all three MPCs use the same horizon (20 steps) and trot gait settings, with 10 steps for stance and 10 steps for leg swing. The MPC time step is adjusted between 0.02, 0.025, and 0.03~\si{\second} to make swing times of 0.2, 0.25, and 0.3~\si{\second}, respectively. The same WBIC setup (e.g., parameters, tasks, constraints) is used across all three MPCs, with identical swing trajectory and landing location selection. Thus, the observed differences in balance control performance primarily result from differences in MPC formulation.

We also tested a full-body dynamics-based MPC, introduced in \cite{khazoom2024tailoring}, to assess whether incorporating a full dynamics model could further enhance locomotion capabilities. While this approach demonstrated stable walking with slow stepping, its computation time was significantly longer (over 25~\si{\milli\second} on a laptop with a 2.4 GHz 8-Core Intel Core i9, whereas ours complete the computation in 1~\si{\milli\second}) and made swing leg control more challenging. We acknowledge that further optimization could reduce computation time and enable smoother swing leg control and touchdown. However, given the limited computational power of the onboard embedded PC and the potential difficulty in incorporating terrain height, we opted for our single rigid body model-based MPC, which we implemented and tested on the physical robot.

\subsection{Hardware Experiments}
\label{subsec:hw}

To test our controller on hardware, we built a standalone robot system based on the Unitree Go1 robot~\cite{unitree}. To run our MPC + WBIC controller on board, an embedded computer -- Beelink SEi12 PC, equipped with a 12th Gen i7-12450H CPU -- is mounted on the robot, along with a 24V to 19V buck converter that supplies power directly from the robot to the PC. Our MPC horizon is set to 24 steps with a time step of 0.026~\si{\second}, resulting in a 0.286~\si{\second} swing phase in a trot gait.

To quantitatively evaluate the noise reduction achieved by our controller relative to the default controller, we measured the sound level generated by the robot while varying walking speeds. A researcher holding a smartphone at ear level walked alongside the robot to record audio, as shown in Fig.~\ref{fig:hardware_exp} (a). The robot walked in a straight line and passed through a designated two-meter noise recording zone, during which sound data were collected. In addition, a camera recorded this zone to verify the robot's entry and exit times, ensuring accurate validation of its walking speed.

\begin{figure}
    \centering
    \includegraphics[width=\linewidth]{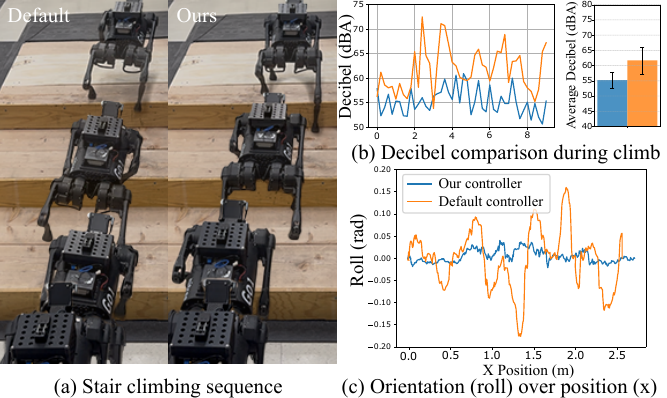}
    \caption{{\bf Stair Climbing Tests.} Comparison of the default and proposed controllers on a 12.7 cm rise, 60 cm tread stair-climbing task, demonstrating our controller improves stability, reduces roll, and lowers noise.}  
    \label{fig:stair}
\end{figure}

We compared the noise levels generated by the default controller and our proposed controller at four different speeds ($0.6\si{\meter\per\second}$, $0.8\si{\meter\per\second}$, $1.0\si{\meter\per\second}$, and $1.2\si{\meter\per\second}$). As shown in Fig.\ref{fig:hardware_exp} (c), our controller reduces noise by nearly 10 dB compared to the default controller, demonstrating a substantial improvement in noise suppression. The noise level obtained in our experiment (50\si{\decibel} on average) is lower than the noise level reported in \cite{valsecchi2024accurate}, which tested their RL-based controller on the Anymal robot. While their focus was being energy-efficient rather than noise suppression, this difference is noteworthy, as a heavier robot with a high center of mass has slower dynamics, which benefits slow gait stabilization. Moreover, 50~\si{\decibel} is even lower than the noise level of wheeled systems tested in \cite{wang2023can} (65~\si{\decibel}). This is a particularly meaningful accomplishment since the participants in \cite{wang2023can} preferred wheeled systems over quadruped robots due to their high noise levels.

\begin{figure}[t]
  \centering
  \includegraphics[width=\linewidth]{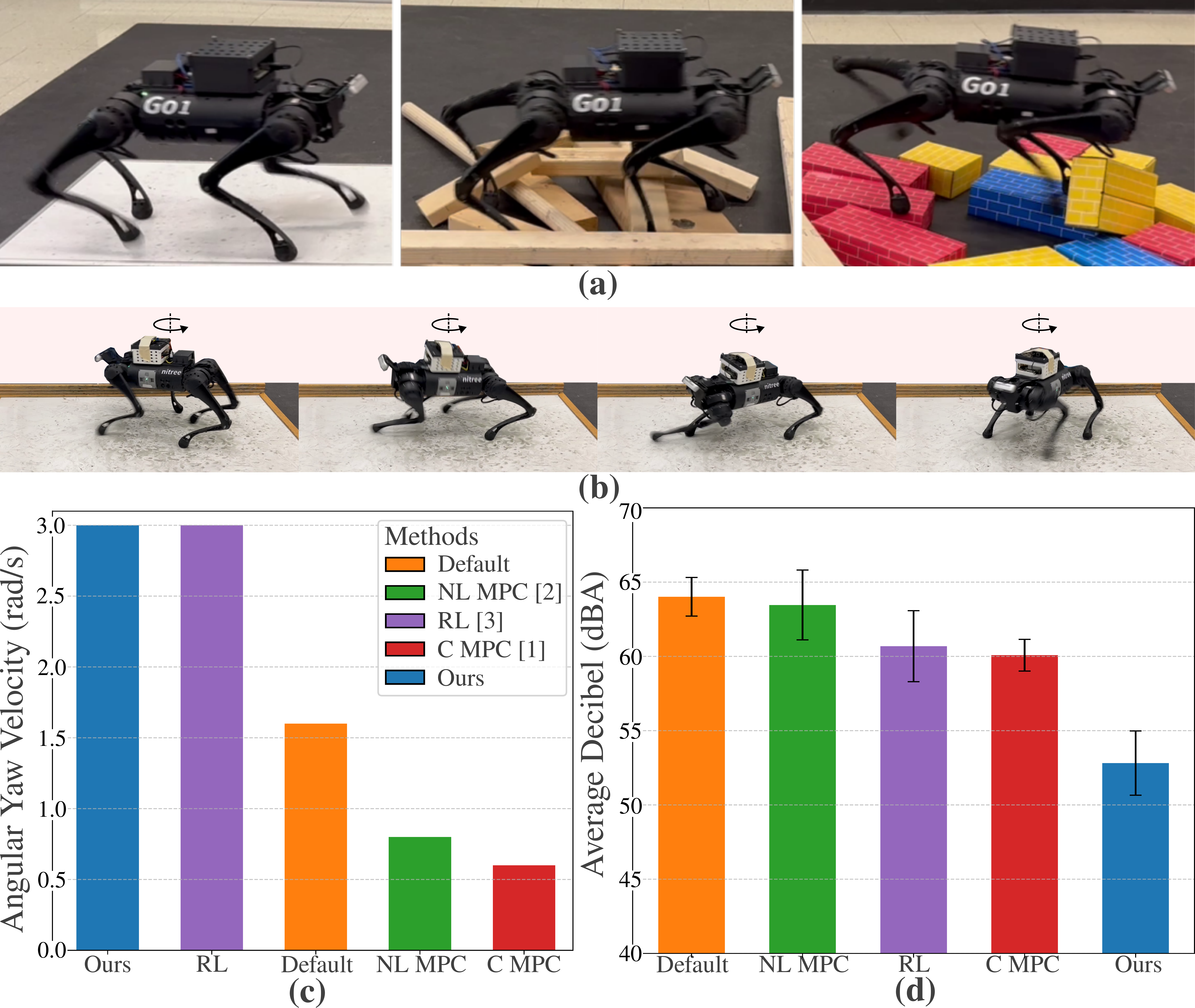}
  \caption{\textbf{Comparison Experiments}. (a) Our method is tested across various challenging terrains, including a low-friction slippery terrain made by a soapy water mixture on a whiteboard, rigid wood blocks of varying dimensions, and a pile of cardboard boxes, demonstrating its robustness and stability. (b) Thanks to the robustness of our controller, the robot can recover balance after slips on the slippery terrain. (c) Our controller outperformed all other optimization-based methods and is comparable to RL, known for excelling on slippery surfaces. (d) Additional noise tests showed our method remains the quietest among all methods. 
  \label{fig:robustness}}
  \vspace{-0.2cm}
\end{figure}

Since our controller employs a slower gait frequency, each step covers a greater distance than the default controller. As shown in Fig.\ref{fig:hardware_exp} (b), when both controllers walk at $1.0~\si{\meter\per\second}$, our controller travels approximately $0.32~\si{\meter}$ in a single step, whereas the default controller spans about $0.12~\si{\meter}$. Despite the longer body movement during the extended two-contact stance period, our method exhibits superior balance control. Fig.\ref{fig:hardware_exp} (d) shows that the robot's body orientation remains well-maintained throughout walking, as the pitch and roll angular velocities for both our controller and the default controller remain around zero. In the case of linear motion, our controller showed even superior performance over the default controller by showing smaller linear velocity oscillations as presented in Fig.\ref{fig:hardware_exp} (e). Not only is the amplitude reduced, but the resultant motion of our controller also demonstrates significantly lower-frequency movement, whereas the default controller’s high-frequency stepping induces excessive vibration, amplifying noise and compromising smoothness. Our controller’s smoother velocity profile results in lower noise and a user experience more similar to that of a guide dog, as also noted by a participants in our user study.

We also evaluated our controller on a stair-climbing task and compared its performance to the default controller. Each stair had a $13~\si{\centi\meter}$ rise and a $60~\si{\centi\meter}$ tread depth, making it shorter and wider than standard stairs. We fabricated this staircase specifically for the stair-climbing test using the Go1 robot, as its current leg kinematics are not sufficiently long to safely traverse standard stairs with an $18~\si{\centi\meter}$ rise and a $28~\si{\centi\meter}$ run. To perceive terrain height, a Realsense D435 was mounted at the front of the robot to capture depth (point cloud) data. For terrain representation, we generated a 2.5D height map using Elevation Mapping~\cite{Fankhauser2018ProbabilisticTerrainMapping}. Given the elevation map, our controller adapts the foot landing height, resulting in lighter contacts and improved overall stability. As illustrated in Fig.\ref{fig:stair} (a) and (c), the stair-climbing mode of the robot's default controller exhibits excessive roll, whereas our controller achieves superior balance control, resulting in lower noise levels as shown in Fig.\ref{fig:stair} (b).

We conducted additional experiments to evaluate the generalization, robustness, and walking noise of our locomotion controller, comparing its performance against state-of-the-art optimization-based and reinforcement learning (RL) approaches.

\subsubsection{Challenging Terrain Walking}
This experiment tested forward walking over diverse challenging terrains, including a whiteboard coated with soapy water, wooden blocks of varying dimensions, and a pile of cardboard boxes (see Fig.~\ref{fig:robustness} (a)). Across ten trials, our controller successfully traversed all surfaces without a single failure, highlighting its strong robustness in unstructured environments.

\subsubsection{Turning on Slippery Surfaces}
To evaluate the stability of our controller, we gradually increased the robot’s yaw velocity on a slippery board until failure occurred. Despite frequent slips (see Fig.~\ref{fig:robustness} (b)), the robot maintained balance and successfully recovered at yaw angular velocities exceeding 3.0$\text{rad/s}$. We compared our controller against leading optimization-based methods - Convex MPC~\cite{kim2019highly} and Nonlinear MPC~\cite{grandia2023perceptive} - as well as an RL-based method with a noise reduction reward, as described by Kim et al.~\cite{kim2025understanding}. For consistency, all controllers operated at a gait frequency of approximately 2$\text{Hz}$ (except the default controller due to limited low-level code access). Our controller outperformed optimization-based methods in maximum stable yaw velocity and matched the robustness of the RL method (see Fig.~\ref{fig:robustness} (c)).

\subsubsection{Acoustic Noise Evaluation}
We quantified acoustic noise during straight-line walking on flat terrain, following the previous procedure. As illustrated in Fig.~\ref{fig:robustness} (d), our controller consistently produced substantially lower noise levels compared to all baseline methods.

\begin{table}
  \caption{User Study Participant Demographics}
  \label{tab:participants}
  \begin{tabular}{cccccc}
    \toprule
    \textbf{ID} & \textbf{Age} & \textbf{Gender} & \textbf{Vision Level} & \textbf{Aid} & \textbf{\makecell[c]{Experience\\(GD yrs$^{*}$ / Robot$^{\dag}$)}}\\
    \midrule
    RH01 & 63 & F & Totally blind & Dog & 33 / \cmark  \\
    RH02 & 66 & F & Legally blind & Dog & 9 / \xmark  \\
    RH03 & 66 & M & Totally blind & Dog & 13 / \cmark  \\
    RH04 & 72 & F & Legally blind & Dog & 29 / \cmark  \\
    
    \bottomrule
\end{tabular}
\footnotesize{\\$^{*}$ Indicates the number of total years the participant worked with guide dogs.}
\footnotesize{\\$^{\dag}$ Denotes the experience of walking with a guide dog robot prior to this study.}\\
\end{table}

\section{User Study}
In this section, we present our user study design and findings, which received approval from the Institutional Review Board at the university. We employed a mixed-methods approach to evaluate whether the walking and stair-climbing experience enabled by our controller effectively addresses the challenges identified in our earlier exploratory research. The specific research questions that we investigated are as follows:

\begin{itemize}
    \item \textbf{RQ1:} Does the reduced noise level of our controller make a noticeable difference for BLV individuals, and how does this improvement influence satisfaction, usability, and workload -- particularly at their preferred walking speed -- when compared to an off-the-shelf locomotion controller?
    \item \textbf{RQ2:} How acceptable is our controller’s stair-climbing functionality in terms of stability, movement, and logistics from a BLV user perspective?
    \item \textbf{RQ3:} In comparison to an animal guide dog, what aspects of our guide dog robot need refinement to support real-world deployment?
\end{itemize}

\subsection{Participants}
We recruited four BLV participants from our earlier exploratory study, ensuring they met the following inclusion criteria: (1) age 18 or older, and (2) visual acuity equivalent to or less than legal blindness (20/200), as defined under the Social Security Act \S 1614~\cite{ssa20}. All participants worked with guide dogs as their primary mobility aid. Their experience levels ranged from 9 to 33 years and three participants have experience with robotic assistive systems or a quadruped robot. Demographic information for these participants is provided in Table~\ref{tab:participants}. 

\subsection{Procedure}

Our study consists of four sequential phases designed to systematically assess participants’ experiences and interactions with the guide dog robot:
\subsubsection{Walking with Primary Mobility Aid}
\begin{figure}
    \centering
    \includegraphics[width=\linewidth]{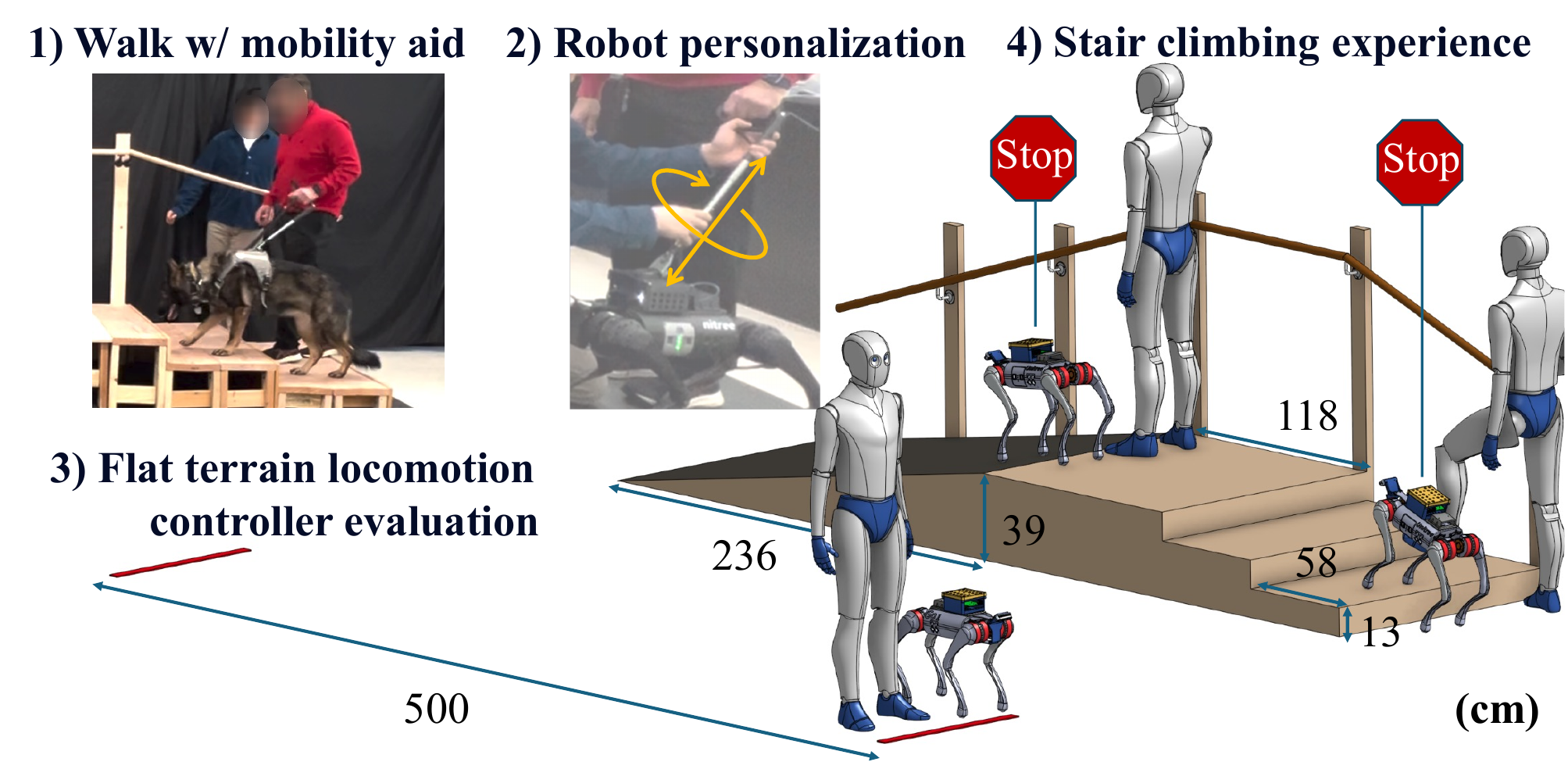}
    \caption{{\bf User Study Procedure and Setup.} The study consists of four sequential phases and is conducted in our lab space, which is equipped with a custom-designed structure featuring a staircase and a ramp. We finalized this setup based on feedback from a guide dog handler regarding the test environment before conducting the user study.}  
    \label{fig:user-study}
\end{figure}

First, participants navigated the test area with their primary mobility aids (in this study, all of them were their guide dogs) alongside one researcher. This is for familiarizing themselves with the environment to perform the same tasks later with the robot: walking in a straight line, climbing a staircase, and descending a ramp as illustrated in Fig.~\ref{fig:user-study}. Afterward, they verbally filled out a survey measuring workload, noise perception, and usability as listed in Section~\ref{subsec:quals}. We proceeded to the next phase only after the participant confirmed their comfort with the environment.

\subsubsection{Robot Familiarization and Personalization}
Next, we introduced the participants to the robot and the harness handle, allowing them to familiarize themselves with handling and positioning. We then customized the robot’s handle length, angle, and forward walking speed to match each participant’s natural gait. At this stage, we employed our locomotion controller (to be evaluated immediately afterward) and had participants perform short walking trials to ensure safety and comfort. Unlike prior studies that used a ``wizard-of-oz'' approach~\cite{kim2025understanding}, we programmed the robot to walk at a fixed velocity to avoid inconsistencies that could arise from human remote control. Moreover, because our controller maintains stable locomotion even at normal walking speeds ($\geq$\SI{1}{\meter\per\second}), participants could freely choose a pace that matched their preference. Once they indicated readiness, we proceeded to the main evaluation of our locomotion controllers.

\subsubsection{Flat Terrain Locomotion Controller Evaluation}
We compared two locomotion controllers: the default controller (Controller A) and our controller (Controller B). To avoid bias, participants were simply informed that they would test ``Controller A'' and ``Controller B.'' Each participant walked with the robot at least three times using A and three times using B, with the order counterbalanced across participants (see Fig.~\ref{fig:user-study}). After testing each controller, participants took part in a semi-structured interview and completed a short survey that recorded ratings for perceived noise, satisfaction, and controller compliance on a 5-point Likert scale, similar to the one in \cite{kim2025understanding}. We also measured workload and usability after each walking session.

\subsubsection{Stair Climbing Experience}
In the last phase, we used our perception-based quiet locomotion controller to guide each participant through a typical guide dog stair-climbing procedure, as detailed in Section~\ref{subsec:stair}. In this test, the robot stopped at the base of the staircase with its front legs on the first step, as depicted in Fig.~\ref{fig:user-study}. Upon receiving a verbal initiation cue from the participant (e.g., ``forward'' or ``go''), a researcher pressed a single `run' button on the remote to command the robot to initiate stair climbing and then stop at the beginning of the down-ramp. A similar initiation process was used for descending the ramp. Like the flat-terrain walking session, the operator was not involved in robot control except for issuing initiation commands. The robot followed a preset, straight-line path at the participant’s chosen speed.



\subsection{Experiment Environment and Robot Hardware}
All trials were conducted in an indoor $5\times5~\si{\meter}$ laboratory space consisting of a flat terrain area, a custom-built staircase, and a ramp (Fig.\ref{fig:user-study}). While the dimensions of the staircase and ramp do not strictly adhere to ADA guidelines due to space constraints and the robot's kinematics limit, the final test environment was designed in consultation with a BLV participant prior to the study to ensure a safe and accessible setup. To maintain consistency across trials, we designated specific floor markings for robot position initialization. Semi-structured interviews and surveys were conducted in a dedicated seating area within the lab to minimize participant fatigue during multiple back-to-back sessions, with at least one break offered. Audio and video data were recorded using two mobile phones and a GoPro for further analysis. The robot used in the trials was the same Go1 robot tested in Section\ref{subsec:hw}, equipped with a rigid harness handle—the same type used for animal guide dogs.

\subsection{Quantitative Metrics}
\label{subsec:quals}
\subsubsection{Workload}
The perceived workload was assessed using the NASA-TLX~\cite{HART1988139}, which measures mental, physical, and temporal demands, along with performance, effort, and frustration. Participants completed the NASA-TLX after each walking session (both flat terrain and stair climbing) for both their primary mobility aid and the robot. 

\subsubsection{Noise}
Perceived noise, compliance, and satisfaction were evaluated after each walking session for the different tasks using a 5-point Likert scale, adapted from \cite{kim2025understanding}. For the evaluation of participants' experience with the robot, we conducted separate surveys for flat-terrain walking and stair climbing. In contrast, for their guide dog, we administered the survey only once to shorten the overall study duration.

\begin{figure}
    \centering
    \includegraphics[width=\linewidth]{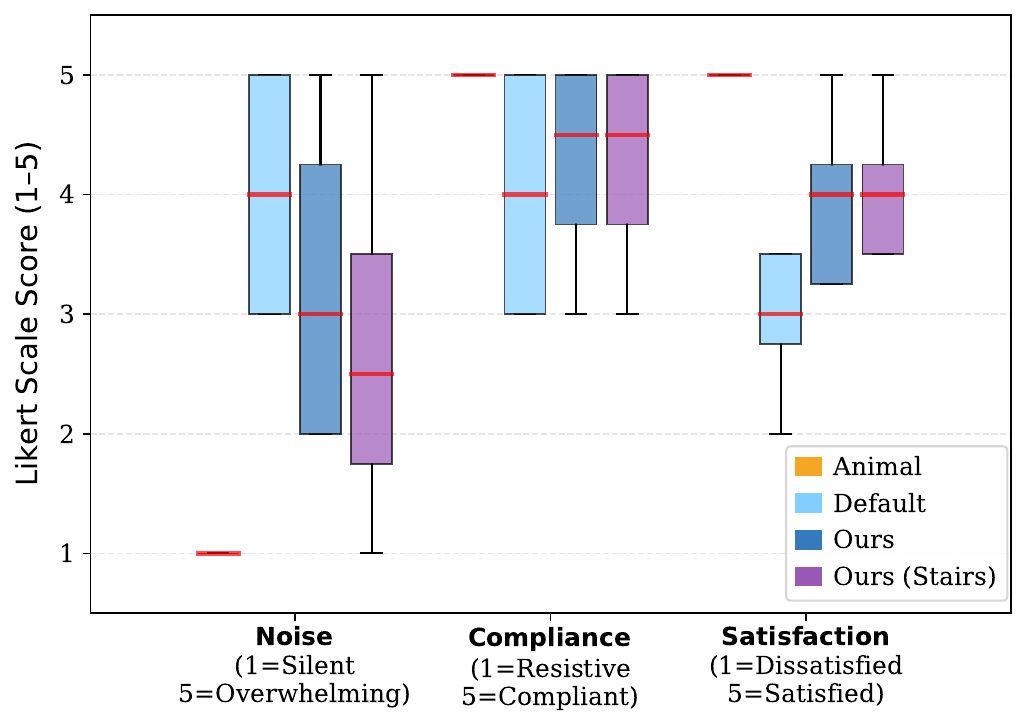}
    \caption{{\bf Scores for noise, compliance, and satisfaction.} Based on the participants' responses, our controller is perceived as quieter and more compliant than the default controller by a noticeable margin. The guide dogs (labeled Animal) received the highest scores in all three categories (noise, compliance, and satisfaction).}  
    \label{fig:noise}
\end{figure}

\subsubsection{Usability}
The System Usability Scale (SUS)~\cite{brooke1996sus} was used to measure the usability after each session in which participants walked on flat terrain using the default locomotion controller, walked on flat terrain using our controller, and climbed stairs using our controller. 


\subsection{Findings}

\subsubsection{Smooth and natural gait with reduced noise}
All participants reported a noticeable reduction in noise when using our locomotion controller compared to the default controller as shown in Fig.~\ref{fig:noise}. Further, we observed that lower noise levels were associated with higher satisfaction, indicated by an inverse relationship between noise and satisfaction (see Fig.~\ref{fig:noise}). 

RH03 noted that the default controller would be disruptive in office or library settings, describing it as loud, similar to a small snowblower. Moreover, RH02 emphasized that such noise can be unsafe for a blind person who tries to listen for signals to cross a street. RH01 emphasized that the default controller's gait was uncomfortable: ``It’s an uncomfortable gait because of the way it kind of jerks up and down much more than the first (ours) gait.'' RH01 also mentioned that our locomotion controller was much smoother, much less noisy, and much more comfortable, offering a more comfortable walking experience overall:


\begin{quote}
{“The first robot (ours) gait was much more natural gait. This one is much more of a forced gait with that bbum bbum bbum bbum. Much more robotic, right? The first robot felt much more, much natural. Much more like a natural gait”} - \textbf{RH01}
\end{quote}

\begin{quote}
{“Second (ours) was better all the way around. Little softer, little less jiggly, and I thought it was smoother.”} - \textbf{RH03}
\end{quote}





\begin{figure}
    \centering
    \includegraphics[width=\linewidth]{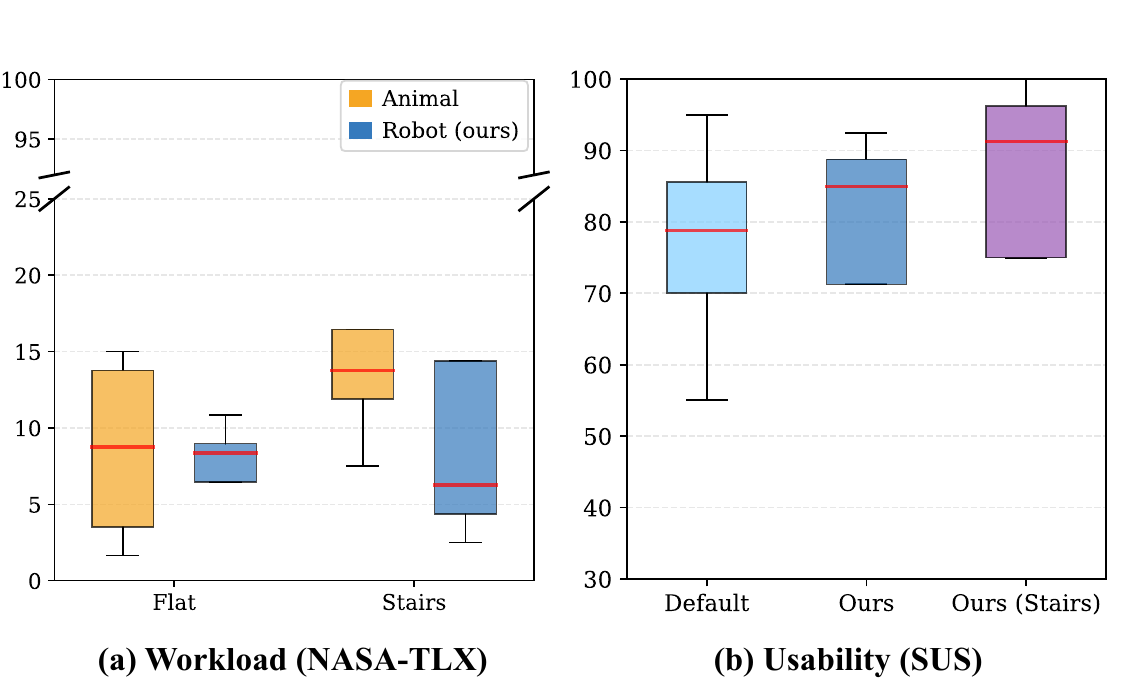}
    \caption{ {\bf Perceived workload and usability scores.} (a) NASA-TLX results reveal a trend of increased workload during stair climbing compared to flat terrain. Participants reported relatively lower workload when using our robot, with the difference being most pronounced in the more demanding stair-climbing condition, (b) System Usability Scale (SUS) scores suggest that our controller may offer improved usability compared to the default, with particularly high ratings observed even during stair climbing.} 
    \label{fig:sus}
\end{figure}

\subsubsection{Comfortable stair climbing}
Participants reported that stair climbing with the robot felt manageable and closely resembled their experiences with animal guide dogs. Very low perceived workload remained consistently low (under 20 on the normalized NASA-TLX) although stair climbing can be more challenging than flat terrain walking as RH02 mentioned (see Fig.~\ref{fig:sus}). Although RH03 initially noted feeling nervous ascending stairs with a robot for the first time, he mentioned that he got comfortable at the third try and even expressed a desire to ascend faster. 

RH01 and RH04 described that stair climbing with the robot was not difficult, but rather similar to that of stepping up with a guide dog. RH04 added that stair climbing was not difficult since she is already working with a guide dog and the experience was very similar. According to RH02, the distinct stopping points made stair climbing easier, and appreciated this feature since guide dogs, although they are trained to stop, can sometimes make mistakes. As one participant summarized, 

\begin{quote}
{``It was very very similar. I like that it gave me time to step up and then I could give it the command when I was ready. I liked that.''} - \textbf{RH04}
\end{quote}

\section{Conclusion}
In this study, we developed a locomotion controller for a guide dog robot specialized in reducing noise. Notably, our work adopts a human-centered approach, including an exploratory study to identify critical problems and clarify the necessary technological developments, system development, and user-centric evaluation.

In the exploratory phase, we met with guide dog handlers and trainers to learn how they interact with guide dogs, the pros and cons of dog guides, and critical perspectives that need to be addressed in robot development. We also experienced blindfolded walking with guide dogs to gain a firsthand understanding of how they assist BLV individuals. Based on these experiences, we concluded that the current quadruped robot controller must be enhanced, especially in terms of noise and vibration suppression.

To achieve dog-like smooth and natural locomotion, we reformulated the MPC framework using a real-time SQP framework, enabling high-speed feedback updates while maintaining accurate angular motion dynamics. The integration of this new MPC with WBIC significantly improves balance control, allowing slow stepping and gentle touchdown during walking -- even at high walking speeds (e.g., $1.2~\si{\meter\per\second}$). This controller reduces the noise of the robot’s walking by $10~\si{\decibel}$ compared to the Unitree Go1 robot's default controller, effectively halving the perceived noise level.

In addition to our efforts to reduce noise, we also incorporated a perception system into our robot to assist BLV individuals not only on flat terrain but also over non-flat terrains such as stairs. Throughout the initial exploratory study, we identified that the robot needs to convey the sequence of actions to help the handler recognize and prepare for upcoming terrain height variations. This feature was implemented alongside our perception-based locomotion controller.

Our efforts in system development culminated in a final human study, in which four guide dog handlers compared their dogs with our robot using two different controllers. Based on survey responses and semi-structured interview results, we found that BLV individuals recognized the reduction in noise and vibration and appreciated the stair-climbing performance, as observed in low workload ratings, high usability scores, and positive feedback on our controllers.

Beyond the three key contributions mentioned in the Introduction, our work has two additional significances: 1) This study represents one of the few human-centered developments of guide dog robots, encompassing the complete cycle from problem identification, technology development, and user feedback, and 2) The proposed controller demonstrated significantly improved balance capability over existing methods and has the potential for applications requiring gentle and smooth motion in legged robots.

\section{Limitation} 
While our research provides promising insights, several inherent limitations must be acknowledged. Participant bias may have influenced our findings due to the limited diversity in technological familiarity, vision levels, primary mobility aids, and ages within our participant pool. Specifically, our interviews were conducted with guide dog handlers experienced with mobility aids similar in form factor to the robot. Additionally, while our study offers initial insights, the small sample size (n=4) limits the generalizability of our findings. Future studies should include larger and more diverse participants such as cane users and individuals with varying vision levels to provide more comprehensive insights into the technological feasibility and advancement of guide dog robots.

From a noise suppression perspective, despite our efforts to minimize noise, residual hardware backlash continues to introduce acoustic artifacts. Future work could address this by refining mechanical tolerances or implementing improved backlash compensation strategies. Another notable source of noise is the thermal management system, where onboard heat dissipation mechanisms contribute to the overall acoustic output. Transitioning to a fanless cooling design or further optimizing airflow dynamics may significantly reduce the audible noise produced by onboard computing hardware. Regarding stair walking, the current leg design imposes kinematic constraints that limit the robot’s ability to descend stairs effectively. Overcoming this challenge will require re-engineering the leg morphology or adjusting its dimensions to enhance adaptability to complex terrains.

\bibliographystyle{unsrt}
\bibliography{main}

\end{document}